\documentclass[letterpaper]{article} 
\usepackage{aaai24}  
\usepackage{times}  
\usepackage{helvet}  
\usepackage{courier}  
\usepackage[hyphens]{url}  
\usepackage{graphicx} 
\usepackage{natbib}  
\usepackage{caption} 
\usepackage[dvipsnames, svgnames, x11names]{xcolor}
\usepackage{colortbl}
\usepackage{multirow}
\usepackage[ruled]{algorithm2e}
\usepackage{subfigure}
\usepackage{newfloat}
\usepackage{listings}
\DeclareCaptionStyle{ruled}{labelfont=normalfont,labelsep=colon,strut=off} 
\title{ProCC: Progressive Cross-primitive Compatibility for Open-World \leavevmode\\ Compositional Zero-Shot Learning \leavevmode\\ ——Appendix——}
\author{
    Fushuo Huo\textsuperscript{\rm 1}, Wenchao Xu\textsuperscript{\rm 1}\footnote{Corresponding author}, Song Guo\textsuperscript{\rm 3}, Jingcai Guo\textsuperscript{\rm 1, \rm 2}\footnotemark[1], Haozhao Wang\textsuperscript{\rm 4}, Ziming Liu\textsuperscript{\rm 1}, Xiaocheng Lu\textsuperscript{\rm 3}
}
\affiliations{
    \textsuperscript{\rm 1}Department of Computing, The Hong Kong Polytechnic University, Hong Kong SAR\\
    \textsuperscript{\rm 2}The Hong Kong Polytechnic University Shenzhen Research Institute, Shenzhen, China \\
    \textsuperscript{\rm 3}Department of Computer Science and Engineering, The Hong Kong University of Science and Technology, Hong Kong SAR \leavevmode\\
    \textsuperscript{\rm 4}School of Computer Science and Technology, Huazhong University of Science and Technology, Wuhan, China \leavevmode\\
}

\usepackage{bibentry}

\begin{document}
\maketitle

\section{Overview}
The appendix presents more experimental settings, results, and analyses as follows:
\begin{itemize}
\item[1)] We illustrate the details of UT-Zappos, MIT-States, and CGQA datasets and intuitively show the differences between the close-world CZSL and Open-World CZSL (OW-CZSL) settings. We also illustrate the whole
training process in Algorithm 1.
\item[2)] We provide more visual illustrations of Cross-Primitive Compatibility (CPC).
\item[3)] We provide more ablation studies on possible architectures of the proposed network.
\item[4)] We also provide more quantitative analysis, including (1) the training time comparisons, (2) the quantitative experiments of our method in the case of updating the backbone ($\omega$), and (3) quantitative analysis of the performance diversity of object and state classification and the necessity of progressive training strategy.
\item[5)] We provide some visual results of our methods for the qualitative analysis.

\end{itemize}

\begin{algorithm}[t]  
\small
	\caption{Training procedure of ProCC.}
	\LinesNumbered 
	\KwIn{Training data ${D^{s}} = \{ (i,c)|i \in {I^{s}},c \in {C^{s}}\}$, pre-trained $\omega$, learning rate $\lambda_1$, $\lambda_2$, $\lambda_3$}
	\KwOut{Optimal $\varphi_o$, $\varphi_s$, CPC: $\varphi_{o \to s}$, $\varphi_{s \to o}$ }
	\textbf{Initialize:} $\varphi_o$, $\varphi_s$, $\varphi_{o \to s}$, $\varphi_{s \to o}$; 

        \textbf{Stage 1:} \emph{// train $\varphi_o$}

	\While{not converged}{
		Sample a batch from $D^s$ as images ${(i_k)}_{k=1}^n$ with their object labels ${(o_k)_{k=1}^{n}}$ \;
        
        \For{samples in the batch}{
            Compute $\ell_{obj}$ via Equation 3.\;
            Update ${\varphi _o} \leftarrow {\varphi _o} - \lambda_1 {\nabla _{{\varphi _o}}}{\ell_{obj}}$
        }  
	}

         \textbf{Stage 2:} \emph{// train $\varphi_s$ and $\varphi_{o \to s}$}

	\While{not converged}{
		Sample a batch from $D^s$ as images ${(i_k)}_{k=1}^n$ with their state labels ${(s_k)_{k=1}^{n}}$ \;
        
        \For{samples in the batch}{
            Compute $\ell_{state}^{con}$ via Equation 11.\;
            Update ${\varphi _{s \cup o \to s}} \leftarrow {\varphi _{s \cup o \to s}} - \lambda_2 {\nabla _{{\varphi _{s \cup o \to s}}}}{\ell_{state}^{con}}$
        }  
	}

        \textbf{Stage 3:} \emph{// finetune $\varphi_o$, $\varphi_s$, $\varphi_{o \to s}$, and $\varphi_{s \to o}$}

	\While{not converged}{
		Sample a batch from $D^s$ as images ${(i_k)}_{k=1}^n$ with their object and state labels ${(o_k, s_k)_{k=1}^{n}}$ \;
        
        \For{samples in the batch}{
            Compute $\ell_{vp}^{con}$ via Equations 10 and 11.\;
            Update ${\varphi _{total}} \leftarrow {\varphi _{total}} - \lambda_3 {\nabla _{{\varphi _{total}}}}{\ell_{vp}^{con}}$
        }  
	}
 
\end{algorithm}

\section{1. Dataset Details and Training Procedure}

\textbf{UT-Zappos} is a dataset for the shoes and has 50025 images. It contains 12 object classes and 16 state classes, with 83 seen compositions and a total of 192 compositional spaces. \textbf{MIT-States} has 53753 images with 115 state classes and 245 object classes. The seen and all output compositions are 1,262 and 28,175, respectively. \textbf{C-GQA} is the largest dataset that contains 186,577 images with 413 state classes and 674 object classes. It contains 5,592 seen compositions and a full output space of 278,362 compositions, which makes it the most extensive for the OW-CZSL. The detailed statistics of UT-Zappos \cite{utz}, MIT-States \cite{mit}, and C-GQA \cite{le} datasets are listed in Table 1. The output space of closed world setting follows the generalized zero-shot learning \cite{zsl1} that includes seen ($C^s$) and unseen ($C^u$) compositions. Compared with the closed-world setting, the output space of the open-world setting contains full possible compositions without prior knowledge. Concretely, the output space of the OW-CZSL setting is ~5.3, 35, and 153 times that of the closed-world setting for UT-Zappos, MIT-States, and CGQA datasets, respectively, The full possible output space, containing tremendous unfeasible compositions, is the main challenge for the OW-CZSL task. 

We illustrate the training procedure of proposed ProCC in \textbf{Algorithm 1}. All experiments are conducted with NVIDIA RTX3090 GPU on CUDA 11.4 using PyTorch framework.

\section{2. More Visual Illustrations}

\begin{figure}[]
\centering
\includegraphics[width=1.05\columnwidth]{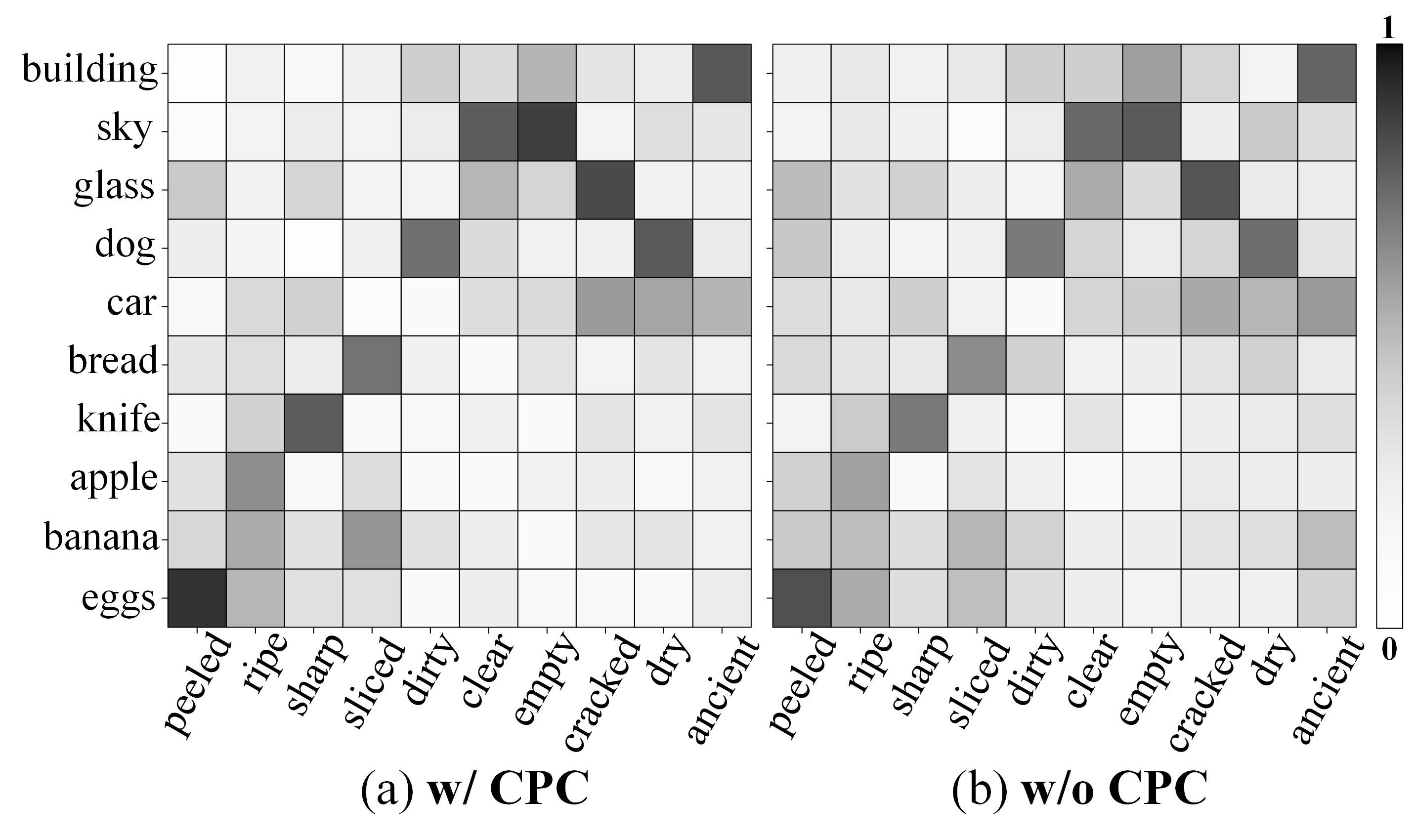}
\vspace{-1em}
\caption{\textbf{Confusion matrices} about prediction probabilities of objects conditioned on states (w/ CPC) or not (w/o CPC).}

\label{fig1}
\vspace{-0.5cm}
\end{figure}

\begin{figure}[]
\centering
\includegraphics[width=0.95\columnwidth]{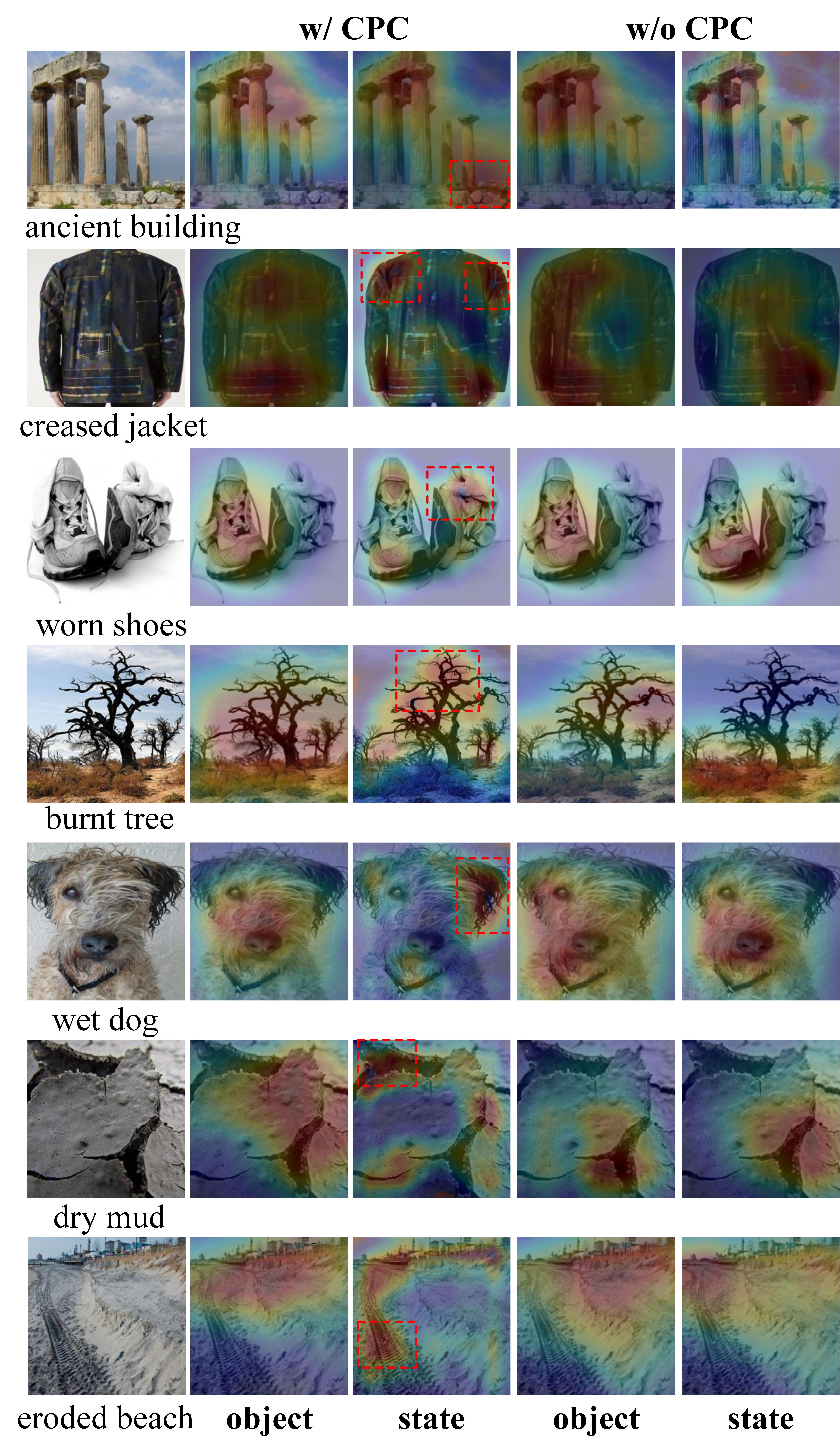}
\caption{\textbf{Visualizations of class activation maps} of ProCC with (w/) and without (w/o) CPC modules on the testing dataset of MIT-States. The discriminative regions are marked with red rectangles.}
\label{fig2}
\end{figure}

To illustrate the effectiveness of the proposed Cross-Primitive Compatibility (CPC) module, we also illustrate the confusion matrices of prediction probabilities of object primitives conditioned on the prediction results of state primitives, as shown in Figure \ref{fig1}. We can learn that the CPC module emphasizes the high-feasibility compositions while eliminating the less-feasibility ones. For example, the possibilities of the $building$ conditioned on the $ancient$ should be higher than the possibilities of the $car$ conditioned on the $ancient$. Because the composition of $ancient$ $car$ is incompatible and the $ancient$ usually describes $building$, $tower$, or $church$ etc., rather than modern objects. Figure \ref{fig2} illustrates the examples of class activation maps (CAM) of seven compositions. The CAMs of states vary greatly as the CPC module drives the network to focus on the critical regions conditioned on objects and related contexts. For instance, as for the $ancient$ $building$ composition, our method can catch the discriminative information about $weathered$ $stones$ (marked in the red rectangle) related to the object $building$. The CAMs of object primitives do not vary much as the prediction of the object usually depends more on the holistic meaning rather than semantic contexts. Based on the visual illustrations, CPC$_{o \rightarrow s}$ brings \emph{more} improvements than CPC$_{s \rightarrow o}$, which is consistent with the quantitative analyses in \textbf{Ablation Study}.

\section{3. More Ablation Studies}

Overall, we follow the Visual Product based methods \cite{rvp, kgsp} and adopt the Multi-Layer Perceptrons (MLP) with three layers as the classifier heads. The learnable Cross-Primitive Memory (CPM) unit consists of $\rm{1d}$ convolution layer and softmax activation function. We conduct the fine-grained analysis of the number of layers in MLP and the kernel size of $\rm{1d}$ convolution layer in CPM. The results are shown in Table 2. We set the number of layers in MLP as 2, 4, and 5, respectively, and also vary the kernel size of $\rm{1d}$ convolution layer in CPM to 1/20, 1/2, 1 of feature dimension. The ablation results are illustrated in Table 2. Firstly, fewer layers in MLP results in poor performance due to losing the ability of non-linear feature extraction. However, the performance also drops to some extent due to overfitting, which corresponds with \cite{rvp}. Then, the CPM with kernel size equal to 1/20 feature dimensions also slightly drops because too few parameters lose the ability to catch the critical information. Meanwhile, cumbersome CPM also leads to the overfitting issue.

\begin{figure}[t]
    \centering
        \subfigure[UT-Zappos]{\includegraphics[width=0.23\textwidth]{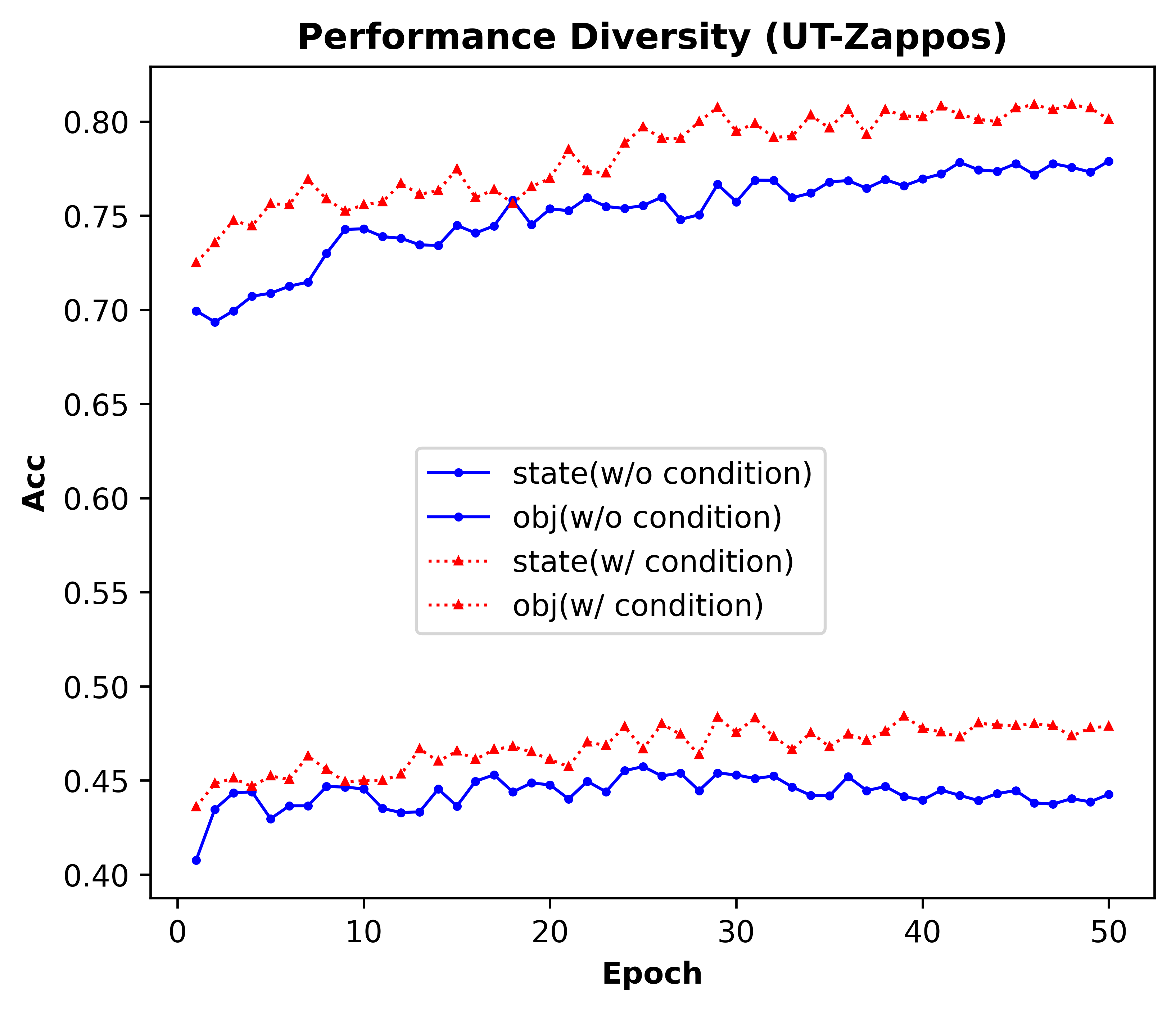}}
        \subfigure[CGQA]{\includegraphics[width=0.23\textwidth]{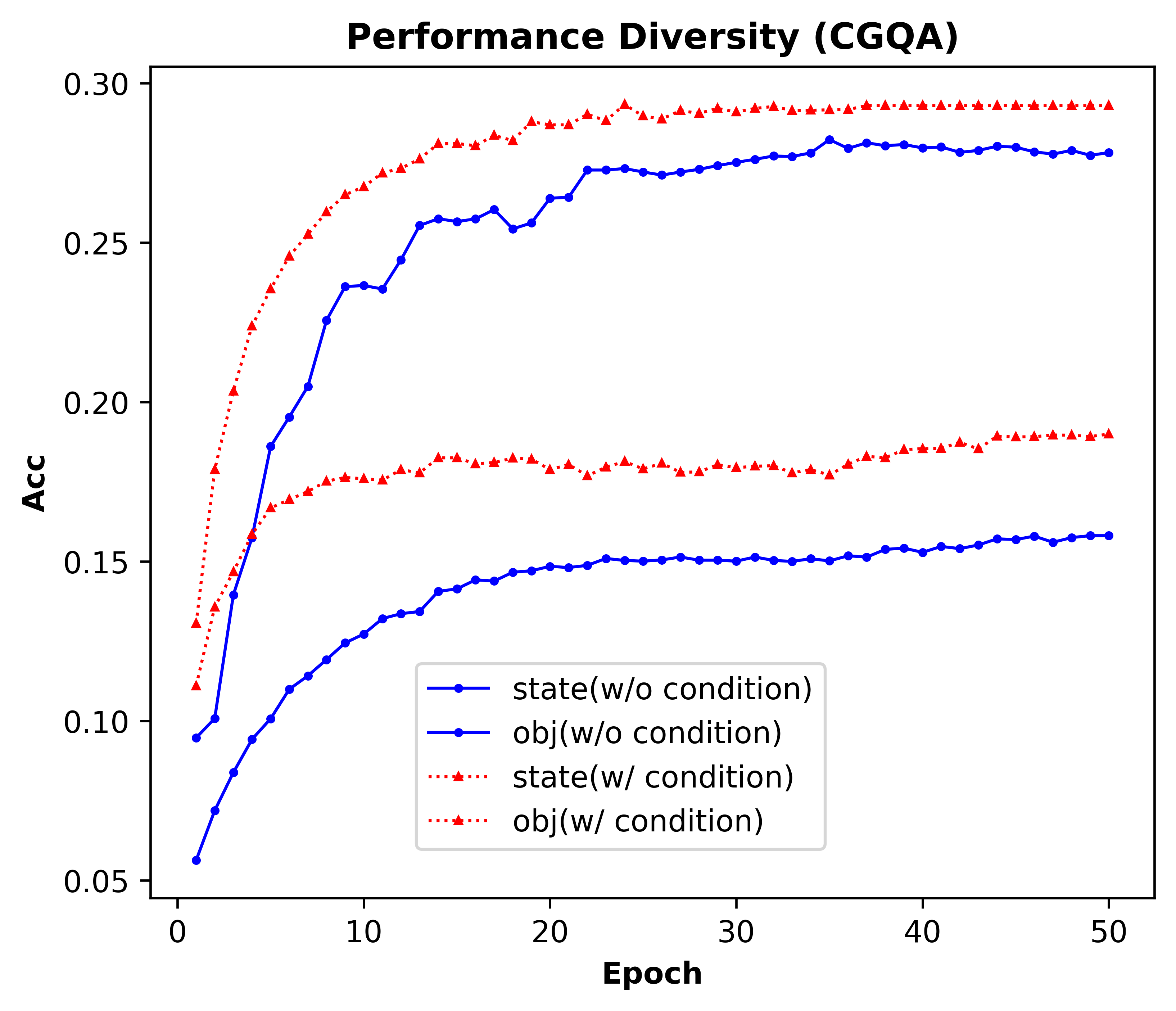}}
        \caption{\textbf{Performance Diversity} of state and object predictions in terms of UT-Zappos and CGQA datasets.}
\end{figure}

\begin{table*}[] \centering
\renewcommand\arraystretch{1.21}
\begin{tabular}{lcc|cc|cccc|cccc}
                                      &     &     & \multicolumn{2}{c|}{Training} & \multicolumn{4}{c|}{Validation}                                  & \multicolumn{4}{c}{Testing}                                     \\ \cline{4-13} 
                                      &     &     &                &              & \multicolumn{2}{c|}{Closed}    & \multicolumn{1}{c|}{Open} &     & \multicolumn{2}{c|}{Closed}   & \multicolumn{1}{c|}{Open} &     \\ \hline
\multicolumn{1}{l|}{\textbf{Dataset}} & s   & o   & $C^s$             & $I^s$           & $C^s$   & \multicolumn{1}{c|}{$C^u$} & \multicolumn{1}{c|}{$C$}    & $I^v$  & $C^s$  & \multicolumn{1}{c|}{$C^u$} & \multicolumn{1}{c|}{$C$}    & $I^t$  \\ \hline
\multicolumn{1}{l|}{UT-Zappos}        & 16  & 12  & 83             & 23k          & 15   & 15                      & 192                       & 3k  & 18  & 18                      & 192                       & 3k  \\
\multicolumn{1}{l|}{MIT-States}       & 115 & 245 & 1262           & 30k          & 300  & 300                     & 28k                     & 10k & 400 & 400                     & 28k                     & 13k \\
\multicolumn{1}{l|}{CGQA}             & 413 & 674 & 5592           & 27k          & 1252 & 1040                    & 278k                      & 7k  & 888 & 923                     & 278k                      & 5k  \\ \hline
\end{tabular}
\caption{The statistics of UT-Zappos \cite{utz}, MIT-States \cite{mit}, and C-GQA \cite{le} datasets. $I$ stands for the image. s and o mean the state and object categories. $C^s$, $C^u$, and $C$ represent the seen, unseen, and full compositions.}
\end{table*}

\begin{table*}[] \centering
\renewcommand\arraystretch{1.2}
\begin{tabular}{l|cccccccc|cccccc}
\multicolumn{1}{c|}{\multirow{3}{*}{\textbf{Method}}} & \multicolumn{8}{c|}{\textbf{OW-CZSL}}                                                                & \multicolumn{6}{c}{\textbf{pCZSL}}                                              \\ \cline{2-15} 
\multicolumn{1}{c|}{}                                 & \multicolumn{4}{c|}{\textbf{CGQA}}                        & \multicolumn{4}{c|}{\textbf{MIT-States}} & \multicolumn{3}{c|}{\textbf{CGQA}}    & \multicolumn{3}{c}{\textbf{MIT-States}} \\ \cline{2-15} 
\multicolumn{1}{c|}{}                                 & \textit{S} & \textit{U} & HM  & \multicolumn{1}{c|}{AUC}  & \textit{S}   & \textit{U}  & HM   & AUC  & S    & U   & \multicolumn{1}{c|}{HM}  & S            & U           & HM         \\ \hline
w/ 2 layers                                           & 27.4       & 1.8        & 3.1 & \multicolumn{1}{c|}{0.42} & 22.8         & 6.8         & 6.4  & 1.0  & 23.4 & 1.0 & \multicolumn{1}{c|}{2.0} & 13.2         & 2.2         & 3.8        \\
w/ 4 layers                                           & 28.4       & 2.6        & 3.5 & \multicolumn{1}{c|}{0.48} & 27.1         & 10.2        & 7.1  & 1.3  & 23.6 & 1.1 & \multicolumn{1}{c|}{2.1} & 13.8         & 2.4         & 4.1        \\
w/ 5 layers                                           & 27.0       & 1.8        & 3.0 & \multicolumn{1}{c|}{0.40} & 22.2         & 7.2         & 6.2  & 0.9  & 23.2 & 1.0 & \multicolumn{1}{c|}{2.0} & 13.4         & 2.4         & 3.8        \\ \hline
w/ 1/20 fd                                            & 27.8       & 2.6        & 3.3 & \multicolumn{1}{c|}{0.44} & 26.5         & 8.9         & 7.0  & 1.4  & 22.9 & 0.9 & \multicolumn{1}{c|}{1.7} & 14.1         & 2.8         & 4.7        \\
w/ 1/2 fd                                             & 28.2       & 2.6        & 3.7 & \multicolumn{1}{c|}{0.51} & 28.0         & 10.3        & 7.6  & 1.5  & 23.1 & 1.1 & \multicolumn{1}{c|}{2.0} & 13.8         & 2.8         & 4.7        \\
w/ 1 fd                                               & 28.0       & 2.4        & 3.6 & \multicolumn{1}{c|}{0.50} & 27.9         & 10.0        & 7.5  & 1.5  & 23.5 & 1.0 & \multicolumn{1}{c|}{2.0} & 13.6         & 2.7         & 4.5        \\ \hline
\textbf{Ours}                                         & 29.0       & 2.6        & 3.8 & \multicolumn{1}{c|}{0.54} & 27.6         & 10.6        & 7.8  & 1.6  & 24.1 & 1.1 & \multicolumn{1}{c|}{2.0} & 14.1         & 2.9         & 4.8        \\ \hline
\end{tabular}
\caption{Ablation studies about the architecture of proposed method in both OW-CZSL and pCZSL settings. fd means the feature dimension.}
\end{table*}

\begin{table}[] \centering
\small
\begin{tabular}{l|cl|cl|cl}
\multicolumn{1}{c|}{\textbf{Method}} & \multicolumn{2}{c|}{\textbf{CGQA}} & \multicolumn{2}{c|}{\textbf{MIT-States}} & \multicolumn{2}{c}{\textbf{UT-Zappos}} \\ \hline
CompCos                              & \multicolumn{2}{c|}{7.8}           & \multicolumn{2}{c|}{3.6}                 & \multicolumn{2}{c}{1.5}                \\
Co-CGE                               & \multicolumn{2}{c|}{21.3}          & \multicolumn{2}{c|}{4.7}                 & \multicolumn{2}{c}{1.6}                \\
KGSP                                 & \multicolumn{2}{c|}{4.1}           & \multicolumn{2}{c|}{2.2}                 & \multicolumn{2}{c}{0.5}                \\
CANet                                  & \multicolumn{2}{c|}{25.6}          & \multicolumn{2}{c|}{11.3}                & \multicolumn{2}{c}{1.1}                \\ \hline
Ours                                 & \multicolumn{2}{c|}{4.7}           & \multicolumn{2}{c|}{2.8}                 & \multicolumn{2}{c}{0.7}                \\ \hline
\end{tabular}
\caption{\textbf{Training time} Comparisons.}
\end{table}

\begin{figure*}[]
\centering
\includegraphics[width=0.95\textwidth]{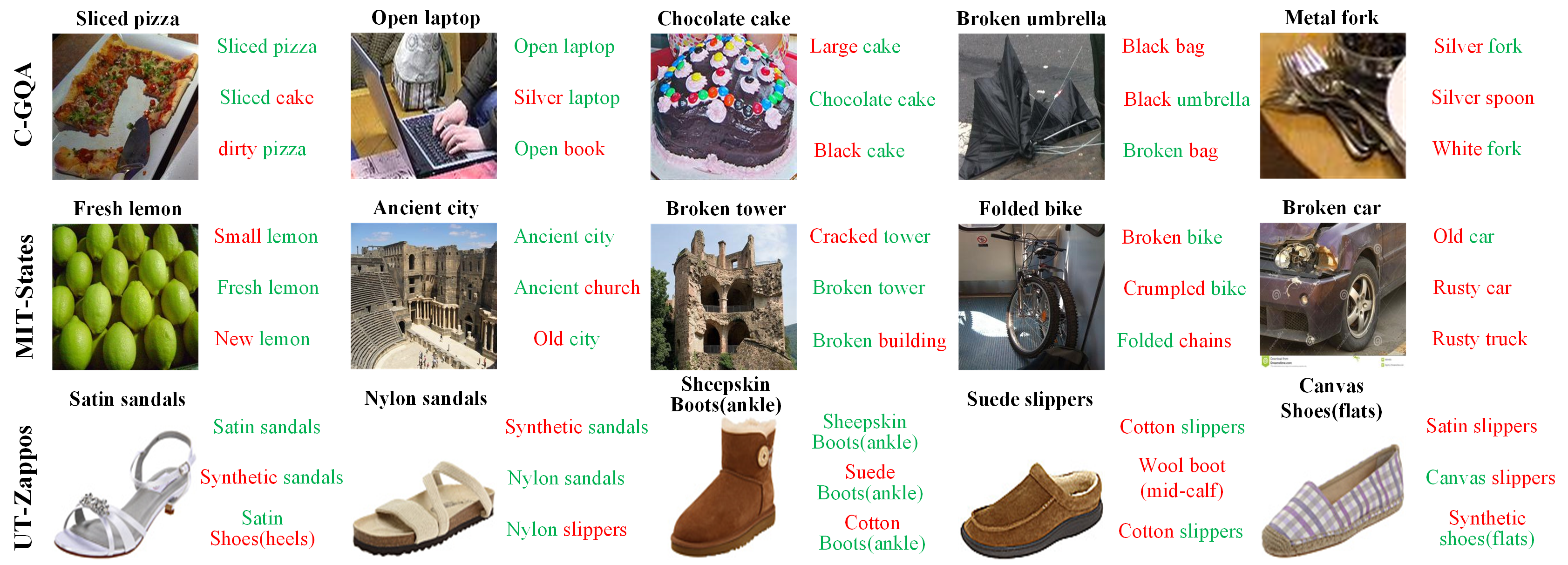}
\caption{Some qualitative results on CGQA, MIT-states, and UT-Zappos datasets. We show the top-3 predictions of our proposed model. The right predictions are marked in green and the wrong results are marked in red.}
\label{fig3}
\end{figure*}

\begin{table*}[]
\renewcommand\arraystretch{1.2}
\setlength{\tabcolsep}{1.0mm}{
\begin{tabular}{l|cccccc|cccccc|cccccc}
\multirow{3}{*}{\textbf{Method}} & \multicolumn{6}{c|}{\textbf{C-GQA}}                                                         & \multicolumn{6}{c|}{\textbf{MIT-States}}                                                   & \multicolumn{6}{c}{\textbf{UT-Zappos}}                                                        \\ \cline{2-19} 
                                 & \multicolumn{2}{c|}{Val}      & \multicolumn{4}{c|}{Test}                                   & \multicolumn{2}{c|}{Val}     & \multicolumn{4}{c|}{Test}                                    & \multicolumn{2}{c|}{Val}       & \multicolumn{4}{c}{Test}                                      \\
                                 & HM            &\multicolumn{1}{c|}{AUC}          & $S$             & $U$            & HM           & AUC           & HM           & \multicolumn{1}{c|}{AUC}          & $S$             & $U$             & HM            & AUC          & HM            & \multicolumn{1}{c|}{AUC}           & $S$             & $U$             & HM            & AUC           \\ \hline
CompCos$_{up}$             & 12.9          & 2.8          & 28.4          & 1.8          & 2.8          & 0.39          & 8.7          & 1.8          & 25.4          & 10.0          & 8.9           & 1.6          & 33.9          & 18.9          & 59.3          & 46.8          & 36.9          & 21.3          \\
CGE$_{up}$               & 12.5          & 2.7          & \underline{ 32.7}    & 1.8          & 2.9          & 0.47          & 8.0          & 1.4          & \textbf{32.4} & 5.1           & 6.0           & 1.0          & 36.0          & 21.7          & 61.7          & 47.7          & 39.0          & 23.1          \\
Co-CGE$_{up}$          & 17.2          & 4.6          & 32.1          & \underline{  3.0}    & \underline{  4.8}    & \underline{  0.78}    & \underline{  8.9}    & \textbf{2.4} & 30.3          & \underline{  11.2}    & \textbf{10.7} & \textbf{2.3} & 38.5          & 22.8          & 61.2          & 45.8          & 40.8          & 23.3          \\
KGSP$_{up}$         & \underline{  17.4}    & \underline{  4.8}    & 31.5          & 2.9          & 4.7          & \underline{  0.78}    & 8.2          & 1.5          & 28.4          & 7.5           & 7.4           & 1.3          & \underline{  39.0}    & \underline{  24.2}    & \underline{  61.8}    & \textbf{52.1} & \underline{  42.3}    & \underline{  26.5}    \\
\textbf{Ours$_{up}$}                             & \textbf{19.2} & \textbf{6.3} & \textbf{33.2} & \textbf{3.2} & \textbf{5.3} & \textbf{0.91} & \textbf{9.2} & \underline{  2.2}    & \underline{  31.9}    & \textbf{11.3} & \textbf{10.7} & \underline{  1.9}    & \textbf{40.9} & \textbf{24.8} & \textbf{64.8} & \underline{  51.5}    & \textbf{43.8} & \textbf{27.9} \\ \hline
\end{tabular}}
\caption{The state-of-the-art comparisons of C-GQA, MIT-States, and UT-Zappos datasets in the OW-CZSL setting. $up$ denotes updating feature extractor ($\omega$). We report the best seen ($S$) and best unseen ($U$) accuracy and best harmonic mean (HM) on the test sub-dataset. $S$, $U$, and HM metrics are also reported on the validation sub-dataset. 
The best and second-best results are \textbf{bold} and {  underlined}}.
\label{table4}
\end{table*}

\begin{table*}[]
\renewcommand\arraystretch{1.2}
\setlength{\tabcolsep}{1.5mm}{
\begin{tabular}{l|cccccc|cccccc|cccccc}
\multirow{3}{*}{\textbf{Method}} & \multicolumn{6}{c|}{\textbf{C-GQA}}                                                        & \multicolumn{6}{c|}{\textbf{MIT-States}}                                                  & \multicolumn{6}{c}{\textbf{UT-Zappos}}                                                      \\ \cline{2-19} 
                                 & \multicolumn{3}{c|}{Val}                   & \multicolumn{3}{c|}{Test}                 & \multicolumn{3}{c|}{Val}                     & \multicolumn{3}{c|}{Test}                 & \multicolumn{3}{c|}{Val}                    & \multicolumn{3}{c}{Test}                     \\
                                 & S             & U           & \multicolumn{1}{c|}{HM}           & S             & U            & HM           & S             & U           & \multicolumn{1}{c|}{HM}           & S             & U            & HM           & S             & U            & \multicolumn{1}{c|}{HM}          & S             & U            & HM            \\ \hline
CompCos$_{up}$                          & 20.3          & 3.4          & 5.8          & 25.9          & 0.7          & 1.4          & 16.7          & 3.2          & 5.4          & 13.4          & 2.2          & 3.8          & 49.1          & 3.8          & 7.0           & 53.6          & 4.0          & 7.4           \\
Co-CGE$_{up}$                           & 23.9          & 4.3          & 7.3          & 22.8          & 0.9          & 1.7          & 17.9          & \underline{  3.5}    & \underline{  5.9}    & 16.1          & \underline{ 2.4 }         & \underline{ 4.2 }         & 54.0          & \underline{  7.1}    &  12.5    & 56.6          & 6.7          & 12.0           \\
KGSP$_{up}$                             & \underline{  27.2}    & \underline{  5.3}    & \underline{  8.9}    & \textbf{26.9}    & \textbf{ 1.2}    & \textbf{2.3}    & \underline{ 19.5} & 3.4          & 5.8    & \underline{  18.4}    &  2.2    &  4.0    & \textbf{ 54.8}    & 7.2          &\underline { 12.7}           & \textbf{ 57.9}    & \underline{  7.4}    & \underline{  13.1}    \\
\textbf{Ours$_{up}$}                             & \textbf{28.0} & \textbf{5.5} & \textbf{9.1} & \underline{  26.2} & \textbf{1.2} & \textbf{2.3} & \textbf{ 20.1}    & \textbf{3.6} & \textbf{6.1} & \textbf{18.9} & \textbf{2.6} & \textbf{4.6} & \underline{ 54.5} & \textbf{7.3} & \textbf{12.9} & \underline{ 57.2} & \textbf{8.0} & \textbf{14.0} \\ \hline
\end{tabular}}
\vspace{-0.5em}
\caption{The state-of-the-art comparisons of C-GQA, MIT-States, and UT-Zappos datasets in the pCZSL setting. $up$ denotes updating feature extractor ($\omega$). We report the seen (S) and unseen (U) accuracy, best harmonic mean (HM) on the testing and validation datasets. 
The best and second-best results are \textbf{bold} and {  underlined}.}
\label{table5}
\end{table*}

\section{4. More Quantitative Analysis}
(1) As the proposed method has a multi-stage training strategy, we provide the training time comparisons in Table 3. All methods are implemented with the same workstation with their default settings. Ours is much more efficient than embedding-based methods \cite{open_pami, open_cvpr, canet}, which also optimize the text embedding to reason the feasible composition embedding. Training KGSP \cite{kgsp} is slightly faster than ours as the KGSP is the baseline of the proposed method with external knowledge.

(2) We also provide more quantitative results in the case of updating the backbone ($\omega$) in Tables \ref{table4} and \ref{table5}. The progressive training strategy imitates the learning strategy to alleviate the negative effect of the partial label setting and the imbalance issue of training multiple tasks. However, updating the shared backbone ($\omega$) causes the forgetting effect \cite{lwf}. The naive solution employs two encoders to respectively extract the features of the object and state primitives or use the continual learning methods \cite{cl_s}. Here, to be fair comparisons, we jointly train the whole network (i.e., only Stage 3). Our method achieves competitive results in terms of the Open-world and partial setting, as shown in Tables 4 and 5.
Concretely, for the OW-CZSL setting, in the largest CGQA dataset, our method consistently outperforms other methods in terms of all metrics. As for the rest two datasets, our method also achieves competitive results compared with other state-of-the-art methods. For the partial label setting, even though without semi-supervised methods, our method also achieves preferable results in terms of all three datasets. The experiments illustrate the importance of the Cross-Primitive Consistency (CPC) module, which excavates discriminative visual attention to achieve feasible state and object compositions.

(3) The conditioned information matters for the CPC module to model the classifiers' interactions. However, classifying states is \emph{more} challenging than objects \cite{dve, kgsp} and joint training inevitably induces noisy conditioned information. Therefore, we propose the progressive training paradigm to alleviate the invalid
cross-primitive interactions based on pre-trained features (also for the pCZSL condition). Here, we provide the quantitative analysis of state and object classifiers with and without pre-trained condition information in Figure 3. We can see that the state and object performance diversity is huge, i.e., classifying states is more challenging than objects and pre-trained condition information eliminates the noisy information and benefits the primitive classifications, especially in the state prediction. Therefore, we design the progressive learning that \emph{firstly} learns to predict object primitives and \emph{then} based on the conditional information from the high-accuracy object classifier, we learn to classify state primitives. \emph{Finally}, based on pre-trained state and object information, we finetune the primitive classifiers and CPC modules.

\section{5. Qualitative Results}
We show some qualitative results for the novel compositions with top-3 predictions in Figure \ref{fig3}. The first three columns show some examples where the top-3 predictions match the ground-truth labels. 
Note that the remaining two predictions of the model can capture at least one factor, which proves that our method can produce high-confident predictions. 

Also, we can see that the last two columns, even failing to recognize the object and state compositions in the top-3 predictions, can also recognize at least one of the state and object primitives. As our method explores the visual attention information merely from the image, the compositions with the same appearance are likely to be confused. For instance, the \emph{broken umbrella} and \emph{broken car} have similar visual appearances to \emph{black bag} and \emph{rusty car}, respectively. However, utilizing the visual information avoids the cumbersome word embedding methods and facilitates real-world applications.

\bibliography{aaai24}

\begin{thebibliography}{12}
\providecommand{\natexlab}[1]{#1}

\bibitem[{De~Lange et~al.(2022)De~Lange, Aljundi, Masana, Parisot, Jia, Leonardis, Slabaugh, and Tuytelaars}]{cl_s}
De~Lange, M.; Aljundi, R.; Masana, M.; Parisot, S.; Jia, X.; Leonardis, A.; Slabaugh, G.; and Tuytelaars, T. 2022.
\newblock A Continual Learning Survey: Defying Forgetting in Classification Tasks.
\newblock \emph{IEEE TPAMI}, 44(7): 3366--3385.

\bibitem[{Isola, Lim, and Adelson(2015)}]{mit}
Isola, P.; Lim, J.~J.; and Adelson, E.~H. 2015.
\newblock Discovering States and Transformations in Image Collections.
\newblock In \emph{CVPR}.

\bibitem[{Karthik, Mancini, and Akata(2021)}]{rvp}
Karthik, S.; Mancini, M.; and Akata, Z. 2021.
\newblock Revisiting Visual Product for Compositional Zero-Shot Learning.
\newblock In \emph{NeurIPS}.

\bibitem[{Karthik, Mancini, and Akata(2022)}]{kgsp}
Karthik, S.; Mancini, M.; and Akata, Z. 2022.
\newblock KG-SP: Knowledge Guided Simple Primitives for Open World Compositional Zero-Shot Learning.
\newblock In \emph{CVPR}, 9336--9345.

\bibitem[{Li and Hoiem(2018)}]{lwf}
Li, Z.; and Hoiem, D. 2018.
\newblock Learning without Forgetting.
\newblock \emph{IEEE TPAMI}, 40(12): 2935--2947.

\bibitem[{Mancini et~al.(2021)Mancini, Naeem, Xian, and Akata}]{open_cvpr}
Mancini, M.; Naeem, M.~F.; Xian, Y.; and Akata, Z. 2021.
\newblock Open World Compositional Zero-Shot Learning.
\newblock In \emph{CVPR}, 5222--5230.

\bibitem[{Mancini et~al.(2022)Mancini, Naeem, Xian, and Akata}]{open_pami}
Mancini, M.; Naeem, M.~F.; Xian, Y.; and Akata, Z. 2022.
\newblock Learning Graph Embeddings for Open World Compositional Zero-Shot Learning.
\newblock \emph{IEEE TPAMI}, 1--1.

\bibitem[{Misra, Gupta, and Hebert(2017)}]{le}
Misra, I.; Gupta, A.; and Hebert, M. 2017.
\newblock From Red Wine to Red Tomato: Composition With Context.
\newblock In \emph{CVPR}.

\bibitem[{Saini, Pham, and Shrivastava(2022)}]{dve}
Saini, N.; Pham, K.; and Shrivastava, A. 2022.
\newblock Disentangling Visual Embeddings for Attributes and Objects.
\newblock In \emph{CVPR}, 13658--13667.

\bibitem[{Wang et~al.(2023)Wang, Liu, Jing, Chen, Liang, Wang, and Shen}]{canet}
Wang, Q.; Liu, L.; Jing, C.; Chen, H.; Liang, G.; Wang, P.; and Shen, C. 2023.
\newblock Learning Conditional Attributes for Compositional Zero-Shot Learning.
\newblock In \emph{CVPR}, 11197--11206.

\bibitem[{Xian et~al.(2019)Xian, Lampert, Schiele, and Akata}]{zsl1}
Xian, Y.; Lampert, C.~H.; Schiele, B.; and Akata, Z. 2019.
\newblock Zero-Shot Learning—A Comprehensive Evaluation of the Good, the Bad and the Ugly.
\newblock \emph{IEEE TPAMI}, 41(9): 2251--2265.

\bibitem[{Yu and Grauman(2014)}]{utz}
Yu, A.; and Grauman, K. 2014.
\newblock Fine-Grained Visual Comparisons with Local Learning.
\newblock In \emph{CVPR}.

\end{thebibliography}

\end{document}